\documentclass[a4paper]{article}

\usepackage{INTERSPEECH2022}

\usepackage{xcolor} % textcolor
\usepackage{algorithm} % algorithm
\usepackage{algpseudocode} % algorithmic
\usepackage{multirow} % multirow/multicol in table
\usepackage{hyperref} % href
\usepackage{cleveref} % cref

\creflabelformat{equation}{#2\textup{#1}#3} % ex) eq. 1

\title{Distilling a Pretrained Language Model to a Multilingual ASR Model}
\name{Kwanghee Choi, Hyung-Min Park}
\address{Sogang University}
\email{juice500@sogang.ac.kr,hpark@sogang.ac.kr}

\begin{document}

\maketitle
\begin{abstract}\label{sec:abstract}
% Unlike large-scale general language models, speech models have not yet shown great generalization performance towards a wide range of tasks, partly because obtaining speech data is more costly.
% To mitigate the problem, we propose Distilling Language model to Speech model (Distill-L2S), which distills the knowledge from a teacher text model to a student speech model, leveraging rich data embedded inside the existing well-trained transformer-based general text model.
% We align the latent representation of two different modalities via our novel knowledge distillation loss, which can be applied to any speech model that yields a sequence-based representations.
% The subtle differences between the text and speech model representations are handled by shrinking mechanism, nearest-neighbor interpolation, and a learnable linear transformation layer.
% We demonstrate the effectiveness of the distillation by applying it to a multilingual automatic speech recognition (ASR) task, where it commonly suffers from the long-tailed distribution of languages.
% We distill the knowledge embedded inside the transformer-based multilingual language model (InfoXLM) to the state-of-the-art ASR model (XLSR-wav2vec 2.0).
% We show the superiority of our method on the CommonVoice dataset, evaluation on total of 18 languages, and provide source codes for the reproduction of the experiments conducted in this paper \footnote{\url{https://anonymous.4open.science/r/xlm_to_xlsr-22D2}}.

Multilingual speech data often suffer from long-tailed language distribution, resulting in performance degradation.
However, multilingual text data is much easier to obtain, yielding a more useful general language model.
Hence, we are motivated to distill the rich knowledge embedded inside a well-trained teacher text model to the student speech model.
We propose a novel method called the Distilling a Language model to a Speech model (Distill-L2S), which aligns the latent representations of two different modalities.
The subtle differences are handled by the shrinking mechanism, nearest-neighbor interpolation, and a learnable linear projection layer.
We demonstrate the effectiveness of our distillation method by applying it to the multilingual automatic speech recognition (ASR) task.
We distill the transformer-based cross-lingual language model (InfoXLM) while fine-tuning the large-scale multilingual ASR model (XLSR-wav2vec 2.0) for each language.
We show the superiority of our method on 20 low-resource languages of the CommonVoice dataset with less than 100 hours of speech data.\footnote{\url{https://github.com/juice500ml/xlm_to_xlsr}}

\end{abstract}
\section{Introduction}\label{sec:intro}

Compared to other machine learning domains, obtaining speech data is expensive, ending up in unavoidable performance degradation of the automatic speech recognition (ASR) model.
Further, collecting non-English speech data is exceptionally difficult as there are not so many speakers available \cite{ardila2020common,winata2020adapt,pratap2020massively}.
In contrast, text models have been successfully developed and widely utilized by improving the performance based on the vast web-crawled corpus \cite{wenzek2020ccnet,chelba2013one,conneau2019unsupervised}.
Especially, the rise of end-to-end (E2E) transformer-based models showed numerous successes \cite{vaswani2017attention,raffel2019exploring,brown2020language,devlin2018bert}, such as cross-lingual language models \cite{conneau2019unsupervised,chi2021infoxlm}.
Motivated by this, transformer architecture is becoming increasingly common for E2E ASR models \cite{winata2020adapt,baevski2021unsupervised,hsu2021hubert,sadhu2021wav2vec,baevski2020wav2vec,schneider2019wav2vec,baevski2019vq}, where its generalization performance is still limited compared to that of text models.
Furthermore, many approaches have been introduced for multilingual ASR, such as adapter modules \cite{winata2020adapt,kannan2019large,hou2021meta}, multi-head architecture \cite{pratap2020massively,sercu2016very,dalmia2018sequence}, logit adjustment \cite{winata2020adapt}, language-dependent batching \cite{kannan2019large,sercu2016very}, multi-task training \cite{toshniwal2018multilingual,li2018multi}, language embeddings \cite{toshniwal2018multilingual,li2018multi}, and cross-lingual training \cite{baevski2021unsupervised}.
However, ASR performance has been limited for minor languages, e.g., languages with less than 100 hours of speech.
Hence, it naturally raises the following intuition: \textit{Can we utilize the well-trained cross-lingual language models to increase the performance of multilingual ASR models on low-resource languages?}
% Despite the advantages of transformers, it is well known that training and deploying large transformer models are costly in terms of computational efficiency.
% Further, despite the advantages of the architecture, it is well known that training and deploying large transformer models are costly in terms of computational efficiency.

One of the main approaches to using one model to help the other is knowledge distillation (KD) \cite{hinton2015distilling,cho2020speech,liu2019end,jiao2019tinybert,sanh2019distilbert,choi2021temporal,kim2021two}.
KD transfers the information embedded inside the latent representations of the teacher model to the student model, which improves the predictive performance of the student model without any architectural modifications \cite{hinton2015distilling}.
Hence, it can enjoy the better of two worlds, easily adding the KD loss in a plug-and-play manner to the existing model training, yet enjoying the better performance with the aid of the teacher model.
For the classification models, it is common to distill by minimizing the distance between two output logits of the teacher and student networks \cite{hinton2015distilling,cho2020speech,sanh2019distilbert}.
Also, many transformer-specific KD methods have been introduced to distill the information more effectively \cite{jiao2019tinybert,sanh2019distilbert,choi2021temporal,kim2021two}.
\cite{jiao2019tinybert} minimizes the distance between the attention maps or feature outputs of two transformer models with different sizes.
% \cite{sun2020mobilebert} uses a more complicated strategy of progressively training the student model by unfreezing only a part of the student model.
\cite{choi2021temporal} distills the attention maps of a transformer-based audio model to smaller architectures such as convolutional or recurrent neural networks.

Although the above methods concentrate on distilling between the same modalities, some methods bridge the differences between speech and text.
\cite{cho2020speech} distills between text and speech models by minimizing the distance between the classification probabilities, whereas \cite{kim2021two} minimizes the distances between the first hidden representations or the output logits.
Both works sidestep the discrepancy problem between speech and text modalities by disregarding the sequence-level features.
\cite{winata2020adapt} handles the multilingual ASR by directly copying the weights of the cross-lingual language model to the ASR model's transformer decoder.
\cite{baevski2021unsupervised} uses $k$-means clustering to segment the audio into phonemic units, where text input is also phonemicized for adversarial training to distinguish the features between two different modalities.
\cite{chung2018unsupervised} aligns the text and speech embeddings via a learnable linear transform layer in an unsupervised manner.
Finally, a simple $n$-gram language model is commonly applied to the final predictions to improve the ASR performance \cite{baevski2021unsupervised,schneider2019wav2vec,conneau2020unsupervised}.
We did not utilize the method in our work, but we emphasize that it can be used in parallel with the KD methods.
KD methods try to yield better predictions in the first place, where using the $n$-gram language model concentrates on refining the output predictions further.

However, aligning the sequence-level temporal features between text and speech data remains challenging.
The shrink mechanism \cite{chen2016phone} is commonly used to align temporal information between speech and text.
It clusters the neighboring outputs into speech segments while discarding the non-speech segments.
\cite{liu2020bridging} propose a model that contains speech and text models as its components, where it concentrates on solving the speech-to-text translation task.
\cite{yi2019ectc} introduces an encoder-decoder architecture with the losses that accommodates each module to solve the ASR task.
\cite{di2019one} inputs speech and text to the transformer encoder and decoder, respectively, improving the performance of speech translation task.

\begin{figure*}[t!] %%% t: top, b: bottom, h: here
\centering
\includegraphics[height=5cm]{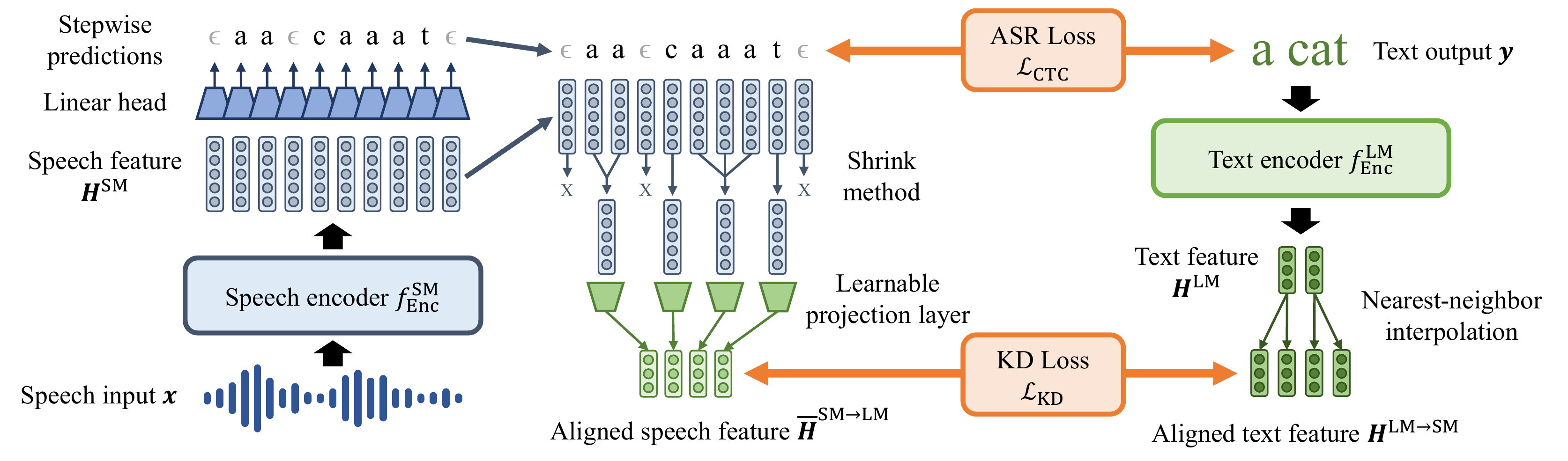}
\caption{
The concept of our proposed method, Distill-L2S.
}
\label{fig:1_model_architecture}
\end{figure*}
In this paper, we propose a novel method, \textit{\textbf{Distill}ing a \textbf{L}anguage model \textbf{to} a \textbf{S}peech model} (Distill-L2S), to leverage the rich knowledge from a well-trained general text model to increase the multilingual ASR performance, especially for low-resource languages.
Our method distills the knowledge by minimizing the differences between text and speech features.
We use various methods to mitigate the inherent discrepancy between the two modalities, such as the shrink method \cite{chen2016phone}, nearest-neighbor interpolation \cite{itseez2014theopencv}, and a learnable linear transformation layer \cite{jiao2019tinybert,chung2018unsupervised}.
Unlike previous methods, our method avoids additional modification of the pre-existing ASR architectures, only requiring the speech features to be sequence-based.
We distill the knowledge of the transformer-based cross-lingual text model, InfoXLM \cite{chi2021infoxlm}, while fine-tuning the transformer-based large-scale ASR model, XLSR-wav2vec 2.0 \cite{conneau2020unsupervised}.
We experiment on 20 low-resource languages of the CommonVoice dataset \cite{ardila2020common} to demonstrate the effectiveness of our method.

\section{Proposed Method: Distill-L2S}\label{sec:method}
In this section, we explain our novel method, \textbf{Distill}ing a \textbf{L}anguage model \textbf{to} a \textbf{S}peech model (Distill-L2S).
We first skim through how ASR models are trained using the CTC loss \cite{graves2006connectionist}.
Then, we describe the speech and text features used for KD.
Finally, we explain our novel loss function that minimizes the distance between features of different modalities.
To assuage the different nature of audio and text modalities, we utilize the shrink mechanism \cite{chen2016phone}, nearest-neighbor interpolation \cite{itseez2014theopencv}, and a learnable linear projection layer \cite{jiao2019tinybert,chung2018unsupervised}.
Our proposed method is illustrated in \Cref{fig:1_model_architecture}.

\subsection{Training ASR models via the CTC Loss}\label{subsec:training_asr}
In this paper, we concentrate on the ASR models that contain hidden states that maintain the sequence-level temporal information.
While numerous architectures can satisfy this condition (e.g., CNNs and RNNs) \cite{choi2021temporal}, we concentrate on the variants of the wav2vec speech model \cite{schneider2019wav2vec}, which utilizes the transformer architecture to achieve state-of-the-art performances \cite{baevski2021unsupervised,hsu2021hubert,sadhu2021wav2vec,baevski2020wav2vec,baevski2019vq}.
We pass raw audio $\mathbf{x}$ to the speech model $f^\text{SM}$ to yield the tokenized output text $\mathbf{y}$:
\begin{align}
    &\text{Input } \mathbf{x}, & &(D_T \times 1)\\
    &\text{Logits } \mathbf{f}_{ij} = f^\text{SM}(\mathbf{x}), & &(D^\text{SM}_{T} \times D^\text{SM}_V)\\
    &\text{Probabilities } \mathbf{p}_{ij} = \texttt{softmax}_j(\mathbf{f}_{ij}), & &(D^\text{SM}_{T} \times D^\text{SM}_V)\\
    &\text{Predictions } \mathbf{\hat{y}} = \texttt{decode}(\mathbf{p}), & &(D^\text{CTC}_{T} \times 1)
\end{align}
where $D_T$ is the length of raw input audio, $D^\text{SM}_{T}$ is the output token length after passing through the speech model $f^\text{SM}$, $D^\text{SM}_{V}$ is the vocabulary size of the speech model's tokenizer, and 
the $\texttt{softmax}_j$ function normalizes the logits $\mathbf{f}$ in the vocabulary axis $D^\text{SM}_{V}$.
We train the model $f^\text{SM}$ using the CTC loss $\mathcal{L}_\text{CTC}$, where $\texttt{decode}$ function denotes the decoding procedure of the CTC loss.
We defer the detailed description to \cite{graves2006connectionist}.

\subsection{Speech Feature Extraction}\label{subsec:speech_feature}
The ASR model $f^\text{SM}$ often yields sequence-level hidden states:
\begin{align}
    &\text{Last hidden state } \mathbf{H}^\text{SM} = f^\text{SM}_\text{Enc}(\mathbf{x}), & &(D^\text{SM}_{T} \times D^\text{SM}_{F})\label{eq:sm_feature}\\
    &\text{Logits } \mathbf{f}_{ij} = f^\text{SM}(\mathbf{x}) = \mathbf{H}^\text{SM} \mathbf{W}_h + \mathbf{b}_h , & &(D^\text{SM}_{T} \times D^\text{SM}_V)
\end{align}
where $D^\text{SM}_{F}$ is the output feature size of the encoder $f^\text{SM}_\text{Enc}$.
$\mathbf{W}_h$ and $\mathbf{b}_h$ are the learnable weights and biases of the final linear head with shape $D^\text{SM}_F \times D^\text{SM}_V$ and $1 \times D^\text{SM}_V$, respectively.
For example, wav2vec \cite{conneau2020unsupervised} feeds the raw audio into multiple 1D convolutional layers, a single linear projection layer, and a transformer encoder to yield the final hidden state.
We employ the last hidden state $\mathbf{H}^\text{SM}$ that contains the sequence-level information for KD.
For brevity, we only describe the linear projection for modeling the logits $\mathbf{f}_{ij}$.
However, we emphasize that our KD loss is not limited to the above case as it only requires the hidden state $\mathbf{H}^\text{SM}$ and the logits $\mathbf{f}_\text{ij}$ to have the same temporal dimension $D^\text{SM}_{T}$.

\subsection{Text Feature Extraction}\label{subsec:text_feature}
Similar to the above, we extract the features for the target text $\mathbf{y}$ using the transformer encoder of the language model $f^\text{LM}_\text{Enc}$:
\begin{align}
    &\text{Tokenized text } \mathbf{y}_\text{tok} = \texttt{tokenize}(\mathbf{y}), &&(D^\text{LM}_{T} \times 1)\\
    &\text{Last hidden state } \mathbf{H}^\text{LM} = f^\text{LM}_\text{Enc}(\mathbf{y}_\text{tok}), &&(D^\text{LM}_{T} \times D^\text{LM}_{F})\label{eq:lm_feature}
\end{align}
where we tokenize the raw text $\mathbf{y}$ using the language model's tokenizer.
$D^\text{LM}_{T}$ is the input text length after tokenization, and $D^\text{LM}_{F}$ is the feature size of the hidden states of the transformer encoder $f^\text{LM}_\text{Enc}$.
We defer further details to \cite{devlin2018bert,chi2021infoxlm}.

\subsection{Feature Alignment}\label{subsec:feature_alignment}
To distill the language model $f^\text{LM}_\text{Enc}$ to the speech model $f^\text{SM}$, we freeze the language model's output features $\mathbf{H}^\text{LM}$ in \cref{eq:lm_feature} and minimize the difference to the speech model's output features $\mathbf{H}^\text{SM}$ in \cref{eq:sm_feature}.
However, both features have largely different temporal and feature dimensions.
We describe the alignment strategies for both perspectives to yield the final KD loss.

\subsubsection{Temporal Dimension Alignment}\label{subsubsec:temporal_dim_alignment}
The inherent difference between speech and text makes temporal alignment difficult.
Models trained with the CTC loss $\mathcal{L}_\text{CTC}$ are known to yield sparse predictions \cite{graves2006connectionist}, making the speech feature have lots of ``empty'' features.
However, tokenized text inputs are fed into the language model; hence there are no ``empty'' text features.
We employ the shrink mechanism \cite{chen2016phone} to mitigate this issue, commonly used to align speech and text features \cite{liu2020bridging,yi2019ectc,di2019one}.
The mechanism has two valuable properties: it removes the ``empty'' information and averages within the same neighboring predictions, i.e., a single segment.
\begin{align}
    &\bar{\mathbf{H}}^\text{SM} = \texttt{shrink}(\mathbf{H}^\text{SM}, \mathbf{p}_{ij}),& &(\bar{D}^\text{SM}_{T} \times D^\text{SM}_{F})
\end{align}
where $\bar{D}^\text{SM}_{T}$ is the temporal dimension after shrinking.

However, the shrunk speech feature $\bar{\mathbf{H}}^\text{SM}$ will still have different temporal dimensions with the corresponding text feature $\mathbf{H}^\text{LM}$.
To fix the issue, we employ a simple nearest-neighbor interpolation method \cite{itseez2014theopencv}, which is commonly used in image processing.
The method naively expands the temporal dimension by duplicating the features multiple times.
As the text tokenizer often has a broader vocabulary (i.e., $D^\text{LM}_V > D^\text{SM}_V$, hence $D^\text{LM}_T < D^\text{SM}_T$), we expand the text feature's temporal dimension to match that of speech:
\begin{align}
    &\mathbf{H}^{\text{LM} \to \text{SM}} = \texttt{interpolate}(\mathbf{H}^\text{LM}).& & (\bar{D}^\text{SM}_{T} \times D^\text{LM}_{F})\label{eq:lm_to_sm}
\end{align}
By applying the two methods to the speech and the text features, both result in the same temporal dimensionx $\bar{D}^\text{SM}_{T}$.

\subsubsection{Feature Dimension Alignment}\label{subsubsec:feat_dim_alignment}
However, the feature dimensions $D^\text{SM}_{F}$ of the speech feature $\bar{\mathbf{H}}^\text{SM}$ are still different from $D^\text{LM}_{F}$ of the text feature $\mathbf{H}^{\text{LM} \to \text{SM}}$.
To fix this, we follow \cite{jiao2019tinybert}, which minimizes the distance between two intermediate representations of the teacher and the student transformer models.
It uses the learnable linear projection layer on one of the representations to translate it to the other.
The distance between the two is minimized via the mean squared error loss $\texttt{MSE}$.
We project the speech feature $\bar{\mathbf{H}}^\text{SM}$ to the text feature $\mathbf{H}^{\text{LM} \to \text{SM}}$ using the linear layer:
\begin{align}
    &\bar{\mathbf{H}}^{\text{SM} \to \text{LM}} = \bar{\mathbf{H}}^\text{SM} \mathbf{W}_f + \mathbf{b}_f, & & (\bar{D}^\text{SM}_{T} \times D^\text{LM}_{F})\label{eq:sm_to_lm}
\end{align}
where $\mathbf{W}_f$ and $\mathbf{b}_f$ is the learnable weights and biases of the projection layer with dimensions $D^\text{SM}_F \times D^\text{LM}_F$ and $1 \times D^\text{LM}_F$, respectively.
Finally, the dimensions of \cref{eq:lm_to_sm} and \ref{eq:sm_to_lm} are successfully matched to minimize the distances between them directly.

\subsection{Knowledge Distillation Loss}\label{subsec:kd_loss}
After the temporal and feature dimension alignment, the text feature $\mathbf{H}^\text{LM}$ and the speech feature $\mathbf{H}^\text{SM}$ are reshaped into $\mathbf{H}^{\text{LM} \to \text{SM}}$ and $\bar{\mathbf{H}}^{\text{SM} \to \text{LM}}$, respectively, ending up in the same shape.
We freeze the teacher language model $f^\text{LM}_\text{Enc}$ and train only the student speech model $f^\text{SM}$ using the loss that jointly optimizes the ASR loss $\mathcal{L}_\text{CTC}$ and the KD loss as follows:
\begin{align}
    &\text{ASR loss } \mathcal{L}_\text{CTC}, \\
    &\text{KD loss } \mathcal{L}_\text{KD} = \texttt{MSE}(\bar{\mathbf{H}}^{\text{SM} \to \text{LM}}, \mathbf{H}^{\text{LM} \to \text{SM}}), \\
    &\text{Final loss } \mathcal{L} = \mathcal{L}_\text{CTC} + \lambda \mathcal{L}_\text{KD},
\end{align}
where $\lambda$ controls the strength of the KD, acting as a trade-off hyperparameter.

\section{Experiments}\label{sec:experiment}
In this section, we verify the effectiveness of our novel method, Distill-L2S.
First, we briefly describe the multilingual ASR task and the CommonVoice dataset \cite{ardila2020common}.
Then, we cover the student speech model and the teacher text language model, utilizing the already pretrained models for both.
We also illustrate the training details for fine-tuning the student speech model on separate languages, where we closely follow the works of \cite{conneau2020unsupervised}.
Finally, similar to \cite{winata2020adapt}, we evaluated the results using the character error rate (CER).
We conduct further analysis of the results.

\subsection{Multilingual ASR}\label{subsec:commonvoice_dataset}
Multilingual speech datasets often encounter the issue of long-tailed language distribution; datasets for minor languages are hard to obtain due to the limited number of speakers \cite{ardila2020common,winata2020adapt,pratap2020massively}.
This is also the case for the CommonVoice dataset \cite{ardila2020common}, a crowdsourced multilingual speech dataset.
Specifically, we used version 6.1, which contains $60$ different languages.
$43$ languages contain $1$ to $100$ hours of speech data.
Our method is ideal for applying to the multilingual ASR for low-resource languages as it can leverage the extensive knowledge of the broader text corpus via the cross-lingual language model.

\begin{table*}[t]
\caption{
CER performance (\%) comparison on the multiple languages when trained on each langauge of the CommonVoice dataset.
% We employ the InfoXLM model \cite{chi2021infoxlm} as the teacher model to distill knowledge to the student model, where the teacher and the student model is trained using the CommonCrawl-100 \cite{chi2021infoxlm,wenzek2020ccnet} and the CommonVoice \cite{ardila2020common} dataset, respectively.
% We report the size of both datasets per language.
% Test character error rate (CER) of the best model found by the validation is reported, where we train the model for $20000$ steps and the validation CER is calculated every $2000$ steps.
The best CER within the same language is written in bold.
}
\label{tab:sota}

% \scriptsize
\centering
\resizebox{0.7\linewidth}{!}{%

\begin{tabular}{c|c|c|c|c|c|ccc}

\toprule
\multicolumn{2}{c|}{Hours of} & GBs of & \multirow{2}{*}{Language} & From scratch & Fine-tuning & \multicolumn{3}{c}{Fine-tuning + Distill-L2S (Ours)} \\
\multicolumn{2}{c|}{Speech} & Text & & $\lambda=0.0$ & $\lambda=0.0$ & $\lambda=0.25$ & $\lambda=0.5$ & $\lambda=1.0$ \\

\midrule
\multirow{5}{*}{ $< 5$h }
& 1 & 14.9 & Czech & 53.07 & 7.40 & 7.43 & \textbf{7.13} & \textbf{7.13} \\
& 1 & 5.9 & Finnish & 73.97 & 11.24 & 11.46 & 12.20 & \textbf{10.87} \\
& 1 & 74.5 & Vietnamese & 88.60 & 32.92 & 33.18 & 33.70 & \textbf{31.99} \\
& 3 & 1.0 & Georgian & 65.15 & 8.83 & 8.87 & 8.92 & \textbf{8.72} \\
& 4 & 2.3 & Lithuanian & 69.15 & 10.99 & \textbf{10.44} & 10.70 & 10.82 \\

\midrule
\multirow{5}{*}{ $< 10$h }
& 5 & 86.8 & Japanese & 100.00 & 100.00 & 100.00 & 100.00 & 100.00 \\
& 7 & 6.2 & Slovenian & 68.23 & 7.57 & 7.50 & \textbf{7.48} & 7.60 \\
& 7 & 1.3 & Latvian & 69.63 & 9.25 & \textbf{8.97} & 9.26 & 9.21 \\
& 8 & 9.5 & Hungarian & 67.61 & 11.49 & 11.99 & 11.59 & \textbf{11.46} \\
& 9 & 11.0 & Romanian & 80.17 & 6.70 & 6.75 & 6.81 & \textbf{6.69} \\

\midrule
\multirow{3}{*}{ $< 20$h }
& 12 & 33.0 & Thai & 85.56 & 13.56 & 13.30 & \textbf{13.27} & 13.80 \\
& 13 & 13.1 & Greek & 69.19 & 10.63 & \textbf{10.33} & 10.42 & 10.62 \\
& 17 & 17.2 & Indonesian & 65.37 & 8.40 & 8.56 & \textbf{8.30} & 8.40 \\

\midrule
\multirow{4}{*}{ $< 50$h }
& 22 & 56.4 & Turkish & 90.27 & 7.59 & \textbf{7.52} & 7.58 & 7.55 \\
& 24 & 7.9 & Tamil & 79.39 & 17.26 & 15.30 & \textbf{15.15} & 15.18 \\
& 27 & 1.4 & Estonian & 51.37 & 8.92 & \textbf{8.34} & 8.71 & 8.48 \\
& 43 & 13.4 & Ukrainian & 59.88 & 8.48 & 8.46 & \textbf{8.44} & 8.50 \\

\midrule
\multirow{3}{*}{ $< 100$h }
& 63 & 39.4 & Portuguese & 59.64 & 5.07 & \textbf{4.98} & 5.07 & 5.09 \\
& 63 & 25.9 & Dutch & 56.45 & 8.33 & 8.72 & 8.62 & \textbf{8.05} \\
& 77 & 16.1 & Arabic & 66.37 & 17.36 & 17.03 & 15.73 & \textbf{14.77} \\

\bottomrule

\end{tabular}

}
\end{table*}

\subsection{Teacher Model: InfoXLM}\label{subsec:teacher_infoxlm}
For the teacher model, we employed InfoXLM \cite{chi2021infoxlm}, which is a cross-lingual transformer-based general language model.
In our experiment, we used the base model that contains $12$ layers and $768$ hidden states.
InfoXLM is pretrained using the recreated CommonCrawl-100 \cite{wenzek2020ccnet}, which contains $94$ languages.
To avoid the performance degradation in the student model-side, we filtered out the languages with less than 1GB of data within CommonCrawl-100, yielding 20 languages for our experiment.
Consult \Cref{tab:sota} for the languages and their dataset sizes.

\subsection{Student Model: XLSR-wav2vec 2.0}\label{subsec:student_xlsr}
Similar to \cite{conneau2020unsupervised}, we fine-tuned XLSR-wav2vec 2.0, a large-scale transformer-based multilingual ASR model.
The model architecture is based on the transformer of $24$ layers and $1024$ hidden states, a total of $317$M parameters.
We chose the pretrained model that is trained on $53$ languages \cite{conneau2020unsupervised}.
The $53$-language dataset is constructed by merging multiple datasets, including the CommonVoice dataset.
We evaluated our models using the test dataset that was not included in the merged training dataset.

\subsection{Training Details}\label{subsec:training_details}
We differed the strength of KD by changing $\lambda \in $ $\{0.0$, $0.25$, $0.5$, $1.0\}$ to monitor the effectiveness of our method.
$\lambda = 0.0$ becomes equivalent to using the CTC loss alone, being the baseline for the other cases.
The transcriptions were tokenized via the unigram algorithm.
We borrowed the settings of \cite{conneau2020unsupervised} with some minor differences.
We evaluated the models per every $2000$ steps on the validation dataset to keep the best model within the total $20000$ training steps.
For the learning rate, we only used the peak learning rate of $2\times 10^{-5}$.
We chose the cosine learning rate schedule with a warm-up of $2000$ steps, slightly different from \cite{conneau2020unsupervised}.
Our experiments were implemented on the \texttt{transformers} library \cite{wolf2020transformers}.
Consult the aforementioned code for further detail.

\subsection{Experimental Results}\label{subsec:exp_results}
\Cref{tab:sota} compares our method on 20 languages with the two baselines, training the model from scratch (weak baseline) and fine-tuning without using our KD method (strong baseline).
All the settings have the same architecture and training details.
Similar to \cite{winata2020adapt}, we did not use the $n$-gram model to remove its influence over the performance and used the CER to evaluate the predictions directly.
Our method wins over the strong baseline on all the language settings except for Japanese, verifying the effectiveness of our novel KD method, Distill-L2S.
We suspect that the number of Japanese tokens ($1249$) is too large for the model to handle.
Using the pretrained cross-lingual model significantly upgrades the weak baseline that utilizes monolingual data only, where our method manages to improve the performance even further.
Especially, Tamil (12\% $\downarrow$) and Arabic (15\% $\downarrow$) show notable decreases in the CER relative to the strong baseline.
However, four languages show less than 1\% CER decrease: Romanian, Hungarian, Ukrainian, and Turkish.

\begin{figure}[t] %%% t: top, b: bottom, h: here
\centering
\caption{
Comparison of the languages with more than $10\%$ (Green) and less than $1\%$ (Red) CER improvements over the corresponding fine-tuning baseline.
}
\includegraphics[width=0.45\textwidth]{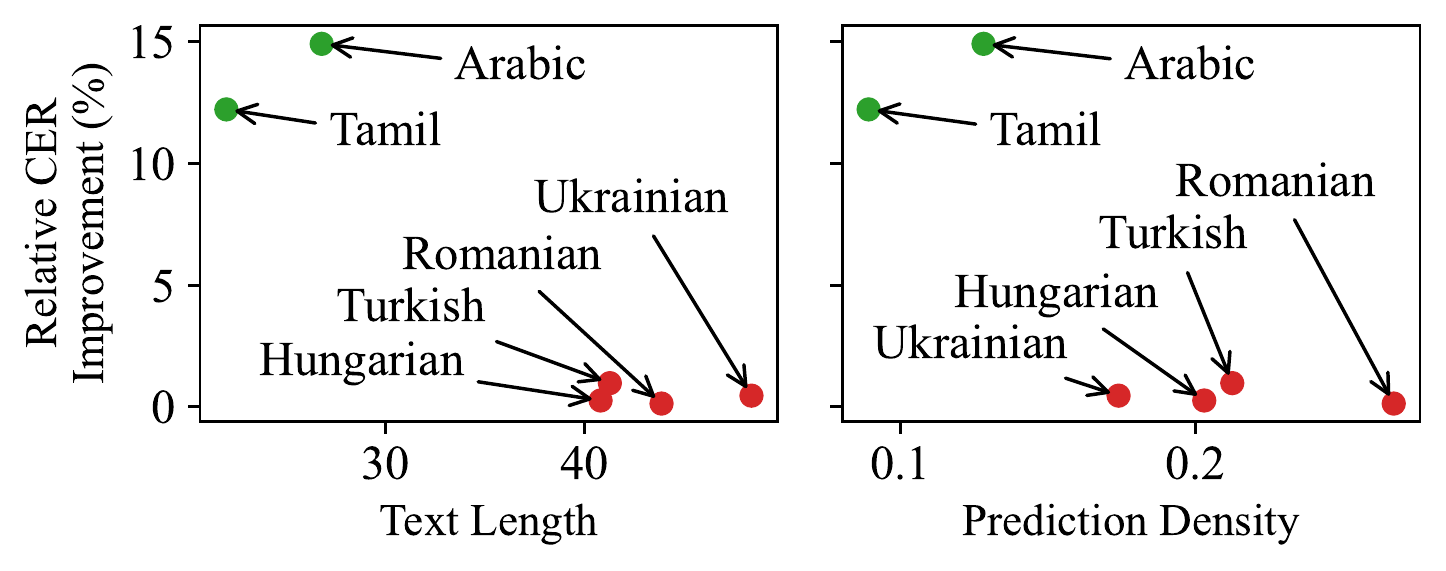}
\label{fig:analysis}
\end{figure}

To analyze why some languages show limited improvement, we compared the two cases above, languages with $>10\%$ or $<1\%$ relative CER improvement in \Cref{fig:analysis}.
We can observe that both the length of the ground truth text and the ratio of the non-blank predictions (Prediction Density) negatively impact our method's performance.
We suspect that as both text and speech features get lengthy, it is more likely for temporal misalignment to occur, showing the limitation of the shrink method and the nearest-neighbor interpolation.
Further work on this issue needs to be done to improve the performance further.

We additionally compare our method with other multilingual ASR methods in \Cref{tab:sota_comparison}.
The methods are as follows: conditioning with one-hot language ID vectors (LID) \cite{li2018multi}, language-specific adapters (LAN) \cite{kannan2019large}, Adapt-and-Adjust (A2) \cite{winata2020adapt}, and the strong baseline with $\lambda=0.0$ (XLSR) \cite{conneau2020unsupervised}.
Notably, A2 also uses a cross-lingual language model, adds language-specific and language-agnostic adapters, and adjusts the logits during inference.
We can see that our method shows superior performance over the compared methods.
\begin{table}[t]
\caption{
CER performance (\%) comparison on Dutch and Turkish.
Columns with \(\dagger\) are the results directly borrowed from \cite{winata2020adapt}.
}
\label{tab:sota_comparison}

% \scriptsize
\centering
\resizebox{0.4\textwidth}{!}{%

\begin{tabular}{c|c|c|c|c|c}

\toprule
\multirow{2}{*}{Language} & LID\(^\dagger\) & LAN\(^\dagger\) & A2\(^\dagger\) & XLSR & Distill-L2S \\
& \cite{li2018multi} & \cite{kannan2019large} & \cite{winata2020adapt} & \cite{conneau2020unsupervised} & (Ours) \\
\midrule
Turkish & 12.0 & 11.8 & 11.3 & 7.59 & \textbf{7.52} \\
Dutch & 16.4 & 16.9 & 15.8 & 8.33 & \textbf{8.05} \\
\bottomrule

\end{tabular}

}
\end{table}

\section{Conclusion}\label{sec:conclusion}
Given that multilingual speech data often suffer from the long-tailed distribution of languages and text data are much easier to obtain, we are motivated to utilize the well-trained cross-lingual language model for the speech task.
Hence, this paper proposes a novel knowledge distillation method, the Distilling a Language model to a Speech model (Distill-L2S), which utilizes the knowledge embedded inside the teacher text model to enhance the predictive performance of the student speech model.
Our method mitigates the discrepancy between features of different modalities by utilizing a set of tricks such as the shrink method, nearest-neighbor interpolation, and the learnable linear projection layer.
We show our method's effectiveness by evaluating in a large number of languages, where our method consistently shows superiority over the compared methods.

\noindent
\textbf{Acknowledgements.}
This work was partly supported by the National Research Foundation of Korea (NRF) grant funded by the Korean government (MSIT) (NRF-2020R1A2B5B01002398) and Institute of Information \& communications Technology Planning \& Evaluation (IITP) grant funded by the Korean government (MSIT) (No. 2021-0-02068, Artificial Intelligence Innovation Hub).

\bibliographystyle{IEEEtran}
\bibliography{mybib}

\begin{thebibliography}{10}
\providecommand{\url}[1]{#1}
\csname url@samestyle\endcsname
\providecommand{\newblock}{\relax}
\providecommand{\bibinfo}[2]{#2}
\providecommand{\BIBentrySTDinterwordspacing}{\spaceskip=0pt\relax}
\providecommand{\BIBentryALTinterwordstretchfactor}{4}
\providecommand{\BIBentryALTinterwordspacing}{\spaceskip=\fontdimen2\font plus
\BIBentryALTinterwordstretchfactor\fontdimen3\font minus
  \fontdimen4\font\relax}
\providecommand{\BIBforeignlanguage}[2]{{%
\expandafter\ifx\csname l@#1\endcsname\relax
\typeout{** WARNING: IEEEtran.bst: No hyphenation pattern has been}%
\typeout{** loaded for the language `#1'. Using the pattern for}%
\typeout{** the default language instead.}%
\else
\language=\csname l@#1\endcsname
\fi
#2}}
\providecommand{\BIBdecl}{\relax}
\BIBdecl

\bibitem{ardila2020common}
R.~Ardila, M.~Branson, K.~Davis, M.~Kohler, J.~Meyer, M.~Henretty, R.~Morais,
  L.~Saunders, F.~M. Tyers, and G.~Weber, ``Common voice: A
  massively-multilingual speech corpus,'' in \emph{LREC}, 2020.

\bibitem{winata2020adapt}
G.~I. Winata, G.~Wang, C.~Xiong, and S.~Hoi, ``Adapt-and-adjust: Overcoming the
  long-tail problem of multilingual speech recognition,'' \emph{Interspeech},
  2021.

\bibitem{pratap2020massively}
V.~Pratap, A.~Sriram, P.~Tomasello, A.~Hannun, V.~Liptchinsky, G.~Synnaeve, and
  R.~Collobert, ``Massively multilingual asr: 50 languages, 1 model, 1 billion
  parameters,'' \emph{Interspeech}, 2020.

\bibitem{wenzek2020ccnet}
G.~Wenzek, M.-A. Lachaux, A.~Conneau, V.~Chaudhary, F.~Guzm{\'a}n, A.~Joulin,
  and {\'E}.~Grave, ``Ccnet: Extracting high quality monolingual datasets from
  web crawl data,'' in \emph{LREC}, 2020.

\bibitem{chelba2013one}
C.~Chelba, T.~Mikolov, M.~Schuster, Q.~Ge, T.~Brants, P.~Koehn, and
  T.~Robinson, ``One billion word benchmark for measuring progress in
  statistical language modeling,'' \emph{Interspeech}, 2014.

\bibitem{conneau2019unsupervised}
A.~Conneau, K.~Khandelwal, N.~Goyal, V.~Chaudhary, G.~Wenzek, F.~Guzm{\'a}n,
  E.~Grave, M.~Ott, L.~Zettlemoyer, and V.~Stoyanov, ``Unsupervised
  cross-lingual representation learning at scale,'' \emph{ACL}, 2020.

\bibitem{vaswani2017attention}
A.~Vaswani, N.~Shazeer, N.~Parmar, J.~Uszkoreit, L.~Jones, A.~N. Gomez,
  {\L}.~Kaiser, and I.~Polosukhin, ``Attention is all you need,''
  \emph{NeurIPS}, vol.~30, 2017.

\bibitem{raffel2019exploring}
C.~Raffel, N.~Shazeer, A.~Roberts, K.~Lee, S.~Narang, M.~Matena, Y.~Zhou,
  W.~Li, and P.~J. Liu, ``Exploring the limits of transfer learning with a
  unified text-to-text transformer,'' \emph{JMLR}, 2020.

\bibitem{brown2020language}
T.~Brown, B.~Mann, N.~Ryder, M.~Subbiah, J.~D. Kaplan, P.~Dhariwal,
  A.~Neelakantan, P.~Shyam, G.~Sastry, A.~Askell \emph{et~al.}, ``Language
  models are few-shot learners,'' \emph{NeurIPS}, 2020.

\bibitem{devlin2018bert}
J.~Devlin, M.-W. Chang, K.~Lee, and K.~Toutanova, ``Bert: Pre-training of deep
  bidirectional transformers for language understanding,'' \emph{NAACL-HLT},
  2018.

\bibitem{chi2021infoxlm}
Z.~Chi, L.~Dong, F.~Wei, N.~Yang, S.~Singhal, W.~Wang, X.~Song, X.-L. Mao,
  H.~Huang, and M.~Zhou, ``Infoxlm: An information-theoretic framework for
  cross-lingual language model pre-training,'' in \emph{NAACL-HLT}, 2021.

\bibitem{baevski2021unsupervised}
A.~Baevski, W.-N. Hsu, A.~Conneau, and M.~Auli, ``Unsupervised speech
  recognition,'' \emph{NeurIPS}, 2021.

\bibitem{hsu2021hubert}
W.-N. Hsu, B.~Bolte, Y.-H.~H. Tsai, K.~Lakhotia, R.~Salakhutdinov, and
  A.~Mohamed, ``Hubert: Self-supervised speech representation learning by
  masked prediction of hidden units,'' \emph{IEEE TASLP}, 2021.

\bibitem{sadhu2021wav2vec}
S.~Sadhu, D.~He, C.-W. Huang, S.~H. Mallidi, M.~Wu, A.~Rastrow, A.~Stolcke,
  J.~Droppo, and R.~Maas, ``Wav2vec-c: A self-supervised model for speech
  representation learning,'' \emph{Interspeech}, 2021.

\bibitem{baevski2020wav2vec}
A.~Baevski, Y.~Zhou, A.~Mohamed, and M.~Auli, ``wav2vec 2.0: A framework for
  self-supervised learning of speech representations,'' \emph{NeurIPS}, 2020.

\bibitem{schneider2019wav2vec}
S.~Schneider, A.~Baevski, R.~Collobert, and M.~Auli, ``wav2vec: Unsupervised
  pre-training for speech recognition,'' \emph{Interspeech}, 2019.

\bibitem{baevski2019vq}
A.~Baevski, S.~Schneider, and M.~Auli, ``vq-wav2vec: Self-supervised learning
  of discrete speech representations,'' \emph{ICLR}, 2019.

\bibitem{kannan2019large}
A.~Kannan, A.~Datta, T.~N. Sainath, E.~Weinstein, B.~Ramabhadran, Y.~Wu,
  A.~Bapna, Z.~Chen, and S.~Lee, ``Large-scale multilingual speech recognition
  with a streaming end-to-end model,'' \emph{Interspeech}, 2019.

\bibitem{hou2021meta}
W.~Hou, Y.~Wang, S.~Gao, and T.~Shinozaki, ``Meta-adapter: Efficient
  cross-lingual adaptation with meta-learning,'' in \emph{ICASSP}, 2021.

\bibitem{sercu2016very}
T.~Sercu, C.~Puhrsch, B.~Kingsbury, and Y.~LeCun, ``Very deep multilingual
  convolutional neural networks for lvcsr,'' in \emph{ICASSP}, 2016.

\bibitem{dalmia2018sequence}
S.~Dalmia, R.~Sanabria, F.~Metze, and A.~W. Black, ``Sequence-based
  multi-lingual low resource speech recognition,'' in \emph{ICASSP}, 2018.

\bibitem{toshniwal2018multilingual}
S.~Toshniwal, T.~N. Sainath, R.~J. Weiss, B.~Li, P.~Moreno, E.~Weinstein, and
  K.~Rao, ``Multilingual speech recognition with a single end-to-end model,''
  in \emph{ICASSP}, 2018.

\bibitem{li2018multi}
B.~Li, T.~N. Sainath, K.~C. Sim, M.~Bacchiani, E.~Weinstein, P.~Nguyen,
  Z.~Chen, Y.~Wu, and K.~Rao, ``Multi-dialect speech recognition with a single
  sequence-to-sequence model,'' in \emph{ICASSP}, 2018.

\bibitem{hinton2015distilling}
G.~Hinton, O.~Vinyals, J.~Dean \emph{et~al.}, ``Distilling the knowledge in a
  neural network,'' \emph{arXiv preprint arXiv:1503.02531}, 2015.

\bibitem{cho2020speech}
W.~I. Cho, D.~Kwak, J.~W. Yoon, and N.~S. Kim, ``Speech to text adaptation:
  Towards an efficient cross-modal distillation,'' \emph{Interspeech}, pp.
  896--900, 2020.

\bibitem{liu2019end}
Y.~Liu, H.~Xiong, Z.~He, J.~Zhang, H.~Wu, H.~Wang, and C.~Zong, ``End-to-end
  speech translation with knowledge distillation,'' \emph{Interspeech}, 2019.

\bibitem{jiao2019tinybert}
X.~Jiao, Y.~Yin, L.~Shang, X.~Jiang, X.~Chen, L.~Li, F.~Wang, and Q.~Liu,
  ``Tinybert: Distilling bert for natural language understanding,'' \emph{EMNLP
  (Findings)}, 2020.

\bibitem{sanh2019distilbert}
V.~Sanh, L.~Debut, J.~Chaumond, and T.~Wolf, ``Distilbert, a distilled version
  of bert: smaller, faster, cheaper and lighter,'' \emph{arXiv preprint
  arXiv:1910.01108}, 2019.

\bibitem{choi2021temporal}
K.~Choi, M.~Kersner, J.~Morton, and B.~Chang, ``Temporal knowledge distillation
  for on-device audio classification,'' \emph{arXiv preprint arXiv:2110.14131},
  2021.

\bibitem{kim2021two}
S.~Kim, G.~Kim, S.~Shin, and S.~Lee, ``Two-stage textual knowledge distillation
  for end-to-end spoken language understanding,'' in \emph{ICASSP}, 2021.

\bibitem{chung2018unsupervised}
Y.-A. Chung, W.-H. Weng, S.~Tong, and J.~Glass, ``Unsupervised cross-modal
  alignment of speech and text embedding spaces,'' \emph{NeurIPS}, 2018.

\bibitem{conneau2020unsupervised}
A.~Conneau, A.~Baevski, R.~Collobert, A.~Mohamed, and M.~Auli, ``Unsupervised
  cross-lingual representation learning for speech recognition,''
  \emph{Interspeech}, 2021.

\bibitem{chen2016phone}
Z.~Chen, Y.~Zhuang, Y.~Qian, and K.~Yu, ``Phone synchronous speech recognition
  with ctc lattices,'' \emph{IEEE TASLP}, 2016.

\bibitem{liu2020bridging}
Y.~Liu, J.~Zhu, J.~Zhang, and C.~Zong, ``Bridging the modality gap for
  speech-to-text translation,'' \emph{arXiv preprint arXiv:2010.14920}, 2020.

\bibitem{yi2019ectc}
C.~Yi, F.~Wang, and B.~Xu, ``Ectc-docd: An end-to-end structure with ctc
  encoder and ocd decoder for speech recognition.'' in \emph{Interspeech},
  2019.

\bibitem{di2019one}
M.~A. Di~Gangi, M.~Negri, and M.~Turchi, ``One-to-many multilingual end-to-end
  speech translation,'' in \emph{IEEE ASRU}, 2019.

\bibitem{itseez2014theopencv}
\emph{The OpenCV Reference Manual}, 2nd~ed., Itseez, 2014.

\bibitem{graves2006connectionist}
A.~Graves, S.~Fern{\'a}ndez, F.~Gomez, and J.~Schmidhuber, ``Connectionist
  temporal classification: labelling unsegmented sequence data with recurrent
  neural networks,'' in \emph{ICML}, 2006.

\bibitem{wolf2020transformers}
T.~Wolf, L.~Debut, V.~Sanh, J.~Chaumond, C.~Delangue, A.~Moi, P.~Cistac,
  T.~Rault, R.~Louf, M.~Funtowicz \emph{et~al.}, ``Transformers:
  State-of-the-art natural language processing,'' in \emph{EMNLP (Demos)},
  2020.

\end{thebibliography}

\end{document}